\documentclass[sn-mathphys-num]{sn-jnl}
\usepackage{graphicx}%
\usepackage{multirow}%
\usepackage{amsmath,amssymb,amsfonts}%
\usepackage{amsthm}%
\usepackage{lmodern}
\usepackage{mathrsfs}%
\usepackage[title]{appendix}%
\usepackage{xcolor}%
\usepackage{soul}
\usepackage{subcaption} 
\usepackage{textcomp}%
\usepackage{booktabs}%
\usepackage{listings}%

\begin{document}

\title {Multistep Brent Oil Price Forecasting with a Multi-Aspect Meta-heuristic Optimization  and Ensemble Deep Learning Model}

\author*[1]{\fnm{Mohammed} \sur{Alruqimi}}\email{mohammed.alruqimi@univr.it}

\author[1]{\fnm{Luca} \sur{Di Persio}}\email{luca.dipersio@univr.it}

\affil*[1]{\orgdiv{Department of Computer Science}, \orgname{University of Verona}, \orgaddress{ \city{Verona}, \country{Italy}}}

\abstract{
Accurate crude oil price forecasting is crucial for various economic activities, including energy trading, risk management, and investment planning. Although deep learning models have emerged as powerful tools for crude oil price forecasting, achieving accurate forecasts remains challenging. Deep learning models' performance is heavily influenced by hyperparameters tuning, and they are expected to perform differently
under various circumstances. Furthermore, price volatility is also sensitive to external factors such as world events. To address these limitations, we propose a hybrid approach that integrates metaheuristic optimization with an ensemble of five widely used neural network architectures for time series forecasting. Unlike existing methods that
apply metaheuristics to optimise hyperparameters within the neural network
architecture, we exploit the GWO metaheuristic optimiser at four levels: feature selection, data preparation, model training, and forecast blending. The proposed approach has been evaluated for forecasting three-ahead days using real-world Brent crude oil price data, and the obtained results demonstrate that the proposed approach improves the forecasting performance measured using various benchmarks, achieving 0.000127 of MSE.}


\keywords{Crude oil price forecasting, Brent oil analysis, time series forecasting, Ensemble learning, Grey Wolf Optimizer}

\maketitle

\section{Introduction}
\label{sec:introduction}
Time series forecasting plays a crucial role in various industries, from finance to logistics to energy, as it provides valuable insights into future trends and helps businesses make informed decisions. However, crude oil forecasting is particularly challenging due to the complex dynamics of the oil market and the non-linear and volatile nature of crude oil prices. 
In recent years, deep learning techniques have gained significant recognition for their ability to capture complex patterns in time series data. Long Short-Term Memory (LSTM) and Gated Recurrent Unit (GRU) networks have become popular choices for modeling temporal dependencies in forecasting tasks \cite{su142114616, article123}.
Despite the emergence of more recent deep learning models such as transformers and generative models, LSTM and GRU remain popular choices for time series forecasting. LSTM and GRU networks excel in tasks with clear temporal sequences \cite{Alruqimi2024, zeng2022transformers}. Moreover, they are more accurate and quicker to train on small datasets, making them well-suited for daily frequency datasets such as daily oil prices \cite{ Alruqimi2024, AysuEzen, Andrade}. To further enhance their performance for time series forecasting, researchers have increasingly integrated additional components, such as CNN-LSTM architectures, combining convolutional neural networks with LSTM to capture spatiotemporal patterns \cite{ALSELWI2024102068}.
Nonetheless, achieving accurate predictions remains a challenge. Individual deep-learning models typically struggle to capture all data patterns, and their performance is heavily dependent on the training configuration. The effectiveness of individual models varies depending on the given dataset and conditions. Furthermore, many external factors influence crude oil prices, including geopolitical events and global economic conditions \cite{2017,ZHAO2023103506, MIAO2017776, FANG2023119329}. 
In other words, the capacity to enhance the capture of dataset patterns, identify external factors influencing the formation of these patterns, and select the appropriate network architecture while configuring it optimally are critical factors that significantly impact forecasting performance. The optimal choice often varies depending on the dataset's characteristics and the specific forecasting problem at hand.
To address these challenges, we propose a hybrid approach to combine the Grey Wolf Optimizer (GWO) with an ensemble of five deep learning networks- LSTM, GRU, CNN-LSTM, CNN-LSTM-attention, and an encoder-decoder-LSTM. GWO is a bio-inspired metaheuristic optimisation algorithm which simulates the leadership hierarchy and grey wolf hunting mechanism in nature. The objective is to make a significant contribution to the advancement of research in the field of crude oil forecasting, building upon the recognition of the challenges mentioned above and the advantages of meta-heuristic optimisation and conditional forecast combination.  This approach can be particularly beneficial in forecasting tasks where different models of diverse structures and data assumptions are expected to perform differently under various conditions. For example, one neural network can perform better than another in a specific dataset and vice-versa.
    
Metaheuristic optimisation techniques present a powerful solution for optimal hyperparameters tuning \cite{Kaveh2023,axioms12030266}. They excels in reaching near-optimal solutions with limited information and computational resources. These algorithms generally give better results and converge faster than blind approaches like Exhaustive Grid Search. This makes them particularly valuable in optimising deep learning methods for time series forecasting.
 While some researchers, such as \cite{ZHANG2021120797}, have already employed metaheuristic optimization algorithms for hyperparameter tuning in deep learning models for crude oil forecasting, we extend this application by utilizing the Grey Wolf Optimizer (GWO) algorithm to optimise the entire process including parameters outside the ML network structures such as feature selection, sliding window, and to fuse multiple forecasts. 
    
The main contributions of this work are summarised as follows:
    \begin{itemize}
        \item An ensemble model to enhance the prediction accuracy of crude oil price. 
        \item Evaluating five different neural network (NN) models with various architectures for multi-step crude oil forecasting. 
        \item Identifying an optimal configuration for each model using the GWO meta-heuristic algorithm.
        \item Introducing a novel method to implement weighted ensemble learning through a meta heuristic optimisation algorithm.
    \end{itemize}

\section{Litrature Review}

Initially, statistical models such as autoregressive integrated moving average (ARIMA) \cite{xiang2013} were applied to forecast oil prices. As statistical methods relies on  mathematical assumptions, they struggle to capture the nonlinear and intricate dynamic of oil prices.  
Recent years have witnessed strong growth in adopting deep learning for time-series forecasting, driven by its remarkable ability to capture more complex patterns within the data. Researchers have employed various deep learning networks, including long short-term memory (LSTM) networks \cite{su142114616,article123, refId0, GUO2023107089}, gated recurrent units (GRUs) \cite{ZHANG2023119617, GUO2023107089, en13071543}, convolutional neural networks (CNNs) \cite{LIN2022109723,wu2023timesnet}, and transformer models \cite{LIU2023e16715}, to forecast crude oil prices accurately based on historical price data and additional data sources such as sentiment analysis, supply and demand data \cite{BECKMANN2020104772, SHANG2021102400}.
A popular approach combines various neural networks layers to improve crude oil forecasting. For instance, many researchers have merged LSTM networks with CNN and self-attention mechanisms to capture temporal, local, and long-term dependencies in historical price data \cite{LIN2022109723, en13071543, ZHANG2023119617, WU2021103357, atmos13040526, QIN2023106266, YANG2021104217}. 
Among these deep learning networks, GRU and LSTM stand out as the most widely applied methods for crude oil price forecasting. These methods excel in capturing temporal dependencies. Additionally, they are well-suited for smaller datasets and are computationally efficient compared to the more complex architectures of Transformers. Readers can refer to \cite{Andrade, ALSELWI2024102068, zeng2022transformers} for further details on this topic. \cite{Alruqimi2024} conducted an empirical analysis of various deep learning architectures, including LSTM, GRU, CNN-LSTM, and Transformer-based models, for oil price forecasting, demonstrating the superiority of GRU and LSTM.
Since tuning model hyperparameters is crucial for enhancing prediction performance, metaheuristic algorithms have recently gained popularity for optimizing the training of machine learning models \cite{Kaveh2023, ABDELBASSET2018185}. These algorithms have been employed to some extent for optimising ML models for crude oil forecasting, i.e. \cite{ZHANG2021120797}, leading to improved prediction accuracy. 
In this work, we present a comprehensive approach to addressing the aforementioned limitations from multiple perspectives by developing an optimized ensemble model based on established models in the literature.

\section{Preliminaries}
    \subsection{Metaheuristic optimisation for machine learning models}
At the core of machine learning models is the process of optimisation, wherein algorithms are trained to perform functions with optimal efficiency. Successful hyperparameter optimisation is essential for achieving precision in a model. The thoughtful selection of model configurations profoundly impacts accuracy and proficiency in handling specific tasks. However, the optimisation of hyperparameters can present a challenging undertaking.
Metaheuristics have become powerful optimisation tools, thanks to their simplicity and capability to effectively address complex NP-hard problems with practical computational resources \cite{oliva2021metaheuristics,CalvetArmasMasipJuan}.
Metaheuristic optimisation algorithms are a family of mathematical-concepts-based methods used to find good solutions to complex optimisation problems. This type of algorithm provides efficient solutions in scenarios where traditional methods may struggle due to complex and dynamic environments. They are designed to explore the search space in a way that is likely to find good solutions, even if they do not always find the optimal solution. Metaheuristics are not bound by specific problem structures, do not need gradient information and can adapt to various optimisation challenges. Their efficiency in solving real-world problems with limited resources and uncertainties makes metaheuristics one of the most notable achievements of the last two decades in operations research 
        \cite{ABDELBASSET2018185, van2023groundwater, van2023mealpy}.
        Metaheuristic algorithms can be grouped into different families based on the natural phenomena they use to guide the search process, such as, for example, evolution or ant behaviour \cite{van2023groundwater}. Over the past few years, population-based methods have been successfully employed to address a diverse range of real-world challenges, including COVID-19 surveillance and forecasting, cloud-edge computing optimisation, energy-efficient cloud computing, feature selection, global optimisation, credit card fraud detection, air pollution forecasting, network intrusion detection, and optimisation of various machine learning models \cite{Kaveh2023}.
        However, the choice of an algorithm largely depends on the optimisation problem at hand. There is no agreed guideline for large-scale nonlinear global optimisation problems for how to choose and what to choose \cite{van2023groundwater}.
            
        \subsection{Grey Wolf Optimizer (GWO)}
\begin{figure}[ht]
                \centering   \includegraphics[]{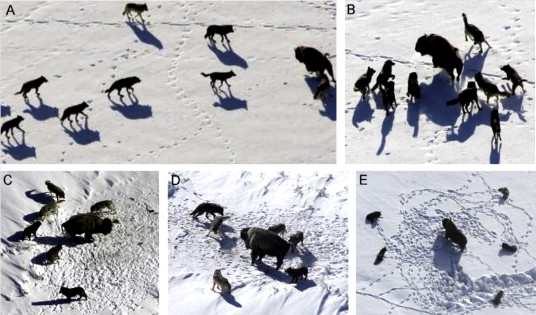}
                \caption{Typical wolf-pack hunting behaviours: (A) chasing, approaching, and tracking prey (B–D) pursuing, harassing, and encircling (E) stationary situation and attack \cite{MURO2011192}.}
                \label{fig:gwo-hunting}
            \end{figure}

            The Grey Wolf Optimizer (GWO) is a bio-inspired swarm intelligence optimisation algorithm inspired by grey wolves, which has been introduced in \cite{MIRJALILI201446} in 2014. This algorithm mimics the leadership hierarchy and hunting mechanism of grey wolves in nature as illustrated in Figure \ref{fig:gwo-hunting}.
 To simulate the leadership hierarchy in the GWO, four types of grey wolves—alpha $(\alpha)$, beta $(\beta)$, delta $(\delta)$, and omega $(\omega)$—are employed. These categories simulate the leadership hierarchy, where the optimisation process is guided by three wolves ($\alpha, \beta $, and  $\delta$), while the omega $(\omega)$ wolves follow. As observed in the hunting process, grey wolves tend to encircle their prey, a phenomenon mathematically modelled by equations (1) and (2).
            
           \begin{align*} \vec{D}=\vert \vec{C}.\vec{X}_{p}(t)-\vec{X}(t)\vert \tag{1}\\ \vec{X}(t+1)=\vec{X}_{p}(t)-\vec{A}.\vec{D} \tag{2} \end{align*}
           Where $t$ is the current iteration, $\vec{A}$  and $\vec{C}$  are coefficient vectors, $\vec{X}_{p}$ is the vector of the prey position, and $\vec{X}$  indicates the vector of the grey wolf position. $\vec{A}$  and $\vec{C}$  vectors can be calculated as illustrated in equations (3) and (4).
           \begin{align*} \vec{A}=2\vec{a}.\vec{r_{1}}-\vec{a} \tag{3}\\ \vec{C}=2.\vec{r_{2}} \tag{4} \end{align*}

           The elements of vector a undergo a linear decrease from 2 to 0 across iterations, with $r_{1}$ and $r_{2}$ being random vectors within the [0, 1] range. In emulating the hunting behaviour of grey wolves, it is assumed that $\alpha$ (the top candidate solution), $\beta$, and $\delta$ possess superior knowledge regarding the potential prey positions. Consequently, the three best solutions obtained thus far are preserved, compelling other search agents (including $\omega$) to adjust their positions based on the optimal search agent positions. Further insights into the GWO algorithm can be found in \cite{MIRJALILI201446, 7560136}.
           
\section{Proposed approach}

In this work, we propose a deep learning ensemble forecasting approach, denoted as GWO-ensemble, to enhance multi-step Brent oil price forecasting. Since hyperparameter tuning is a crucial factor affecting model performance, we integrate the well-known Grey Wolf Optimizer (GWO). Unlike existing models that focus solely on neural network hyperparameter tuning, we apply GWO throughout all stages of model development, including feature selection, sequence data preparation, training individual models, and blending forecasts to generate the final output.

\begin{figure}[htbp]
        \centering {\includegraphics[scale=0.50]{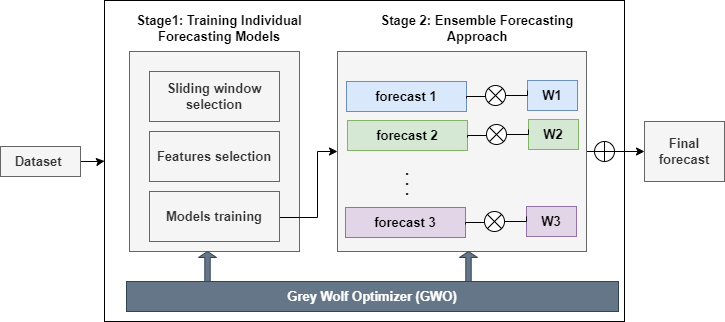}}
        \caption{Proposed Approach}
        \label{fig:Proposed Approach}
    \end{figure}

The proposed approach is implemented in two main stages, as shown in the figure \ref{fig:Proposed Approach}. In the first stage, we train five deep neural networks (described in Section \ref{targeted dl models}), proven for their proficiency in handling time series prediction. In this stage, GWO is used for optimisation of selecting sliding windows for building data sequences, feature selection, and tuning neural network hyperparameters.

The second stage involves blending the forecasts of the individual models using a weight-variant technique. In this stage, GWO is employed to optimize the weights for the forecast blending process.

    To implement the proposed model, we build our code upon mealpy library \cite{van2023mealpy}, where users can select from a wide range of advanced population-based metaheuristic algorithms.

    In Subsection \ref{targeted dl models}, we describe the individual models used as the bases for the ensemble learning. Then, in the next subsection \ref{subsec:weighted_ensemble}, we explain the ensemble learning approach. In Subsection \ref{subsec:gwo-multi-aspect}, we outline aspects and stages in which the GWO optimiser has been employed. 

    \subsection{Individual models}
    \label{targeted dl models}

Our ensemble forecasting model is developed using blended ensemble learning that combines the following five established deep learning  architectures. First, we trained each model individually, and then the individual forecasts obtained by each model were blended using a weighted-variant method, as illustrated in Figure \ref{fig:Proposed Approach}.
        \begin{enumerate}
            \item \textbf{Bi-LSTM}, which is a bidirectional Long Short-Term Memory (Bi-LSTM) model; one LSTM layer followed by two full connected layers with relu activation function. 
            \item \textbf{Bi-GRU}, is a bidirectional Gated Recurrent Unit (Bi-GRU) model; one GRU layer followed by two fully connected layers with a relu activation function. 
            \item \textbf{CNN-Bi-LSTM}, This model combines two convolutional neural networks (CNNs) to extract input features and two LSTM layers for learning temporal dependencies in the data, followed by a fully connected layer with relu activation function.
            \item \textbf{CNN-Bi-LSTM-Attention}, which adds an attention technique for the previous one. 
             \item \textbf{Encoder-decoder-Bi-LSTM} is an encoder-decoder architecture with two LSTMs with 128 units for the encoder and another LSTM with 200 units for the decoder.
        \end{enumerate}

    \subsection{GWO ensemble model}
    \label{subsec:weighted_ensemble}
        Inspired by the advantages of blending forecasts using conditional methods, we explore a weighted ensemble method combined with a meta-heuristic optimiser to fuse forecasts obtained from individual models. The weighted ensemble method in forecasting involves a mathematical approach to combine the predictions of multiple individual models to improve overall accuracy and robustness. Each model in the ensemble contributes to the final forecast with a certain weight. 
        The final ensemble forecast is then obtained by summing the products of each model's forecast with its corresponding optimal weight. Mathematically, this can be represented as the weighted sum:
        
        \begin{align*}
        \text{Weighted ensemble prediction} &= \sum_{i=1}^n M_i W_i \tag{10}\\
        \end{align*}
        Here, $W_i$ represents an optimal weight assigned to  $model_i$, where $0 \leq W_i \leq 1$, and $\sum W_i = 1$. While $M_i$ represents $model_i$ forecasts. 
        In this study, a dynamic weighted ensemble method was employed to blend the predictions of the top five forecasts obtained from the SENT-Bi-GRU, SENT-Bi-LSTM, SENT-CNN-Bi-LSTM, SENT-CNN-Bi-LSTM-att, and SENT-USD-Encoder-Decoder-LSTM. The weights metric $W$ was optimised using the Grey Wolf Optimization (GWO) metaheuristic algorithm.

    \subsection{GWO optimisation} 
    \label{subsec:gwo-multi-aspect}
    
     The Grey Wolf Optimization (GWO) algorithm is employed at four different levels to optimise crucial aspects of the ensemble forecasting process: feature selection, sliding window size selection, model hyperparameter tuning, and final forecast blending. 
     Generally, implementation of metaheuristic optimisations involves defining the main following procedures: problem preparation, objective function definition, population initialisation, and optimisation process. Firstly, the problem preparation involves specifying parameters to be calibrated and establishing their respective ranges (for instance, the range of sliding windows, the list of input features, and the number of epochs, see Table \ref{tab:hyperparameters-config}). The objective function representing the metric that the metaheuristic aims to either maximise or minimise during the optimisation process. This function typically encapsulates the performance of the deep learning model, such as accuracy or loss. The population in the context of metaheuristic optimisation refers to a collection of candidate solutions simultaneously. The metaheuristic explores and refines this population throughout the optimisation iterations to iteratively improve the model's performance.  A larger population generally provides more diversity and allows for a more thorough search space exploration, but it also increases computational cost.
\subsubsection{Feature selection}
The first parameter to be calibrated for each individual model using the GWO is the selection of input features. The range of possible values for this parameter includes four options: [None], [USDX], [SENT], and [USDX, SENT]. This allows the model to be evaluated using only historical Brent oil price data, USDX alone, SENT alone, or a combination of both features.
\subsubsection{Sliding window size}
This parameter represents the temporal scope of historical data used to train a model. Selecting an optimal sliding window size is a delicate balance between incorporating sufficient historical information for meaningful pattern recognition and avoiding overfitting or loss of relevance. Thus, selection suitable sliding window size is crucial to ensure the trained model learns from past data effectively. The defined range for calibrating this parameter spans from 3 to 30, allowing for exploration of the optimal value within this interval.

\subsubsection{Neural networks hyperparameters tuning} Hyperparameter tuning is essential for achieving optimal performance in deep learning models, as it significantly affects training efficiency and accuracy. Finding the right hyperparameters, such as learning rate, optimizer, and hidden units, is challenging due to the complex interplay of parameters and the high-dimensional search space. We applied the GWO algorithm to find the optimal configurations for all the models described in Section \ref{targeted dl models}. Table \ref{tab:hyperparameters-config} presents the optimised hyperparameters and their corresponding range values.

\subsubsection{Blending of final forecasts} Ensemble forecast methods can enhance accuracy by combining forecasts of diverse models that capture different aspects of the data. This often leads to better generalisation and robustness. However, the success of the ensemble methods depends on the individual models' diversity and quality and the nature of the data. Thus, variant-Weight technique is a well-suited method for ensemble forecasting, in which multiple models are combined by assigning distinct weights to their respective predictions \cite{atmos13040526, WU2021103357}. To achieve this, we utilized the GWO algorithm to optimize the weight distribution, thereby determining the contribution of individual models to the final ensemble, as detailed in Section \ref{subsec:weighted_ensemble}.

\section{Experiments}
\label{sec:exp}
    This section outlines the dataset, details the experimental setup, and delves into the obtained results.
    \subsection{Dataset and input features}
    The dataset used for the experiments consists of the daily observations of the Brent crude oil price for the period from (January 3, 2012, to April 1, 2021) including the period where the COVID-19 pandemic has affected energy and stock markets. It also includes Two external factors: the daily USD closing price (USDX) and a daily cumulative sentimental score (SENT).

    Crude oil prices movement is heavily impacted by external factors that shape market dynamics. In our previous work \cite{Alruqimi2024}, we analyzed the correlation between several external variables and Brent oil prices, including stocks, commodities, and sentiment analysis indicators. Our goal was to identify the most valuable variables for improving prediction accuracy while maintaining simplicity and efficiency. Accordingly, In this work, we integrate the U.S. dollar index (USDX) and sentimental index (SENT), individually and combined.
    The evident inverse relationship between Brent crude, which is traded in U.S. dollars, and the U.S. Dollar Index is a valuable factor in modeling Brent prices and is widely utilised in the literature \cite{ZHOU2021126218}. 
    While oil sentiment analysis shows substantial relevance to major events that impact the oil market. Thus, the sentimental index can reflect many other major factors such as supply and demand dynamics. 

\begin{figure}[ht]
        \centering
        \includegraphics[scale=0.60]{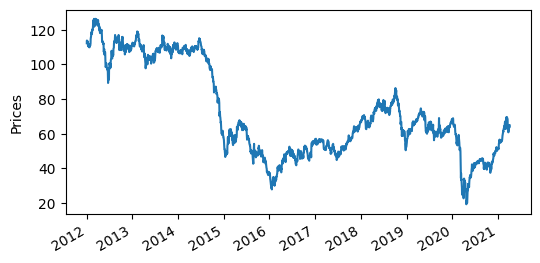}
        \caption{Brent oil price trend from 2012 to 2021}
        \label{fig:main-dataset}
    \end{figure}

    Historical data of Brent oil  and USDX rate can be obtained from many public online sources such as Yahoo Finance. While SENT index is generated by the CrudeBERT model described in \cite{Kaplan2023}. CrudeBERT is a variant of FinBERT that has been fine-tuned towards assessing the impact of market events on crude oil prices, focusing on frequently occurring market events and their impact on market prices according to Adam Smith’s theory of supply and demand. Mainly, CrudeBERT dataset used headlines originating from 1034 unique news sources, published on the Dow Jones newswires (approx. 21,200), Reuters (approx. 3,000), Bloomberg (approx. 1,100), and Platts (approx. 870). The sentimental score data is described and publicly available in \cite{Kaplan2023}. The period from 2012 to 2021 was chosen for analysis due to the availability of sentiment scores during this time frame. Figure \ref{fig:main-dataset} displays the time-series plot of the daily closing price of Brent for this period. 
    
\begin{figure}[!hbt]
        \centering
        \includegraphics[scale=0.50]{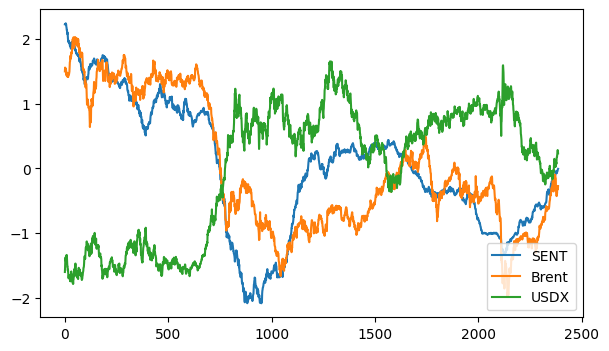}
        \caption{Historical data of Brent oil price, US dollar index (USDX), and the sentimental score (SENT)}
        \label{fig:three-dataset}
    \end{figure}

    Figure \ref{fig:three-dataset} illustrates the trends of Brent price, USD index (USDX), and the sentiment score (SENT) for the period from 2012 to 2021.


    \subsection{Experimental setup}
        \label{sec:problem}

    As mentioned earlier, our experiments are conducted in two stages. In the first stage, the objective is to identify optimal values for hyperparameters of each model (optimal configuration). The individual models are described in  Section \ref{targeted dl models}, and the targeted hyperparameters are described in Table \ref{tab:hyperparameters-config}. Subsequently, we train and evaluate each model individually using its corresponding optimal configuration. The forecasts obtained from each model serve as inputs for the second stage. In the second stage, we aim to optimally blend the forecasts obtained individually from each model in the first stage to produce the final forecast as illustrated in Figure \ref{fig:Proposed Approach}. 

    In order to determine the optimal configuration for each model, the Grey Wolf Optimization (GWO) has been executed through five independent runs for each network. The objective function is specifically designed to minimize Mean Squared Error (MSE) loss, aiming to attain the best global solution. The best global solution is simply a vector containing the optimal values identified for the targeted hyperparameters, corresponding to the smallest Mean Squared Error (MSE) at these values. 

    \begin{table}[htbp!]
        
        \caption{The list of hyperparameters to be optimised by the GWO algorithm for each network, combined with the external features and the window size.  The 'range' column shows the bound of the search space defined for the GWO.}
        \label{tab:hyperparameters-config}
        \setlength{\tabcolsep}{0.5pc}
        \begin{tabular}{@{}lp{2.5cm}p{7.3cm}}
        \toprule
           Parameter     & Range                            & Description \\ \hline
           learning rate & (0.0001 - 0.1)                   & this parameter, crucial for controlling the step size during model training \\ \hline
           hidden unites & ($2^n$, where $1 \leq n \leq 8$) & the number of hidden units \\ \hline
           optimiser    & 7 options                         & The optimisation algorithm employed during training, the seven options are (SGD, RMSprop, Adagrad, Adadelta, AdamW, Adam, and Adamax) \\ \hline
           dropout     & (0.2 - 0.5)         & Dropout is a regularisation technique that randomly drops nodes from the network during training to prevent model overfitting.  \\ \hline
           sliding window size & (3-30) & The number of timesteps used to represent the input data is crucial for capturing the temporal dynamics of the series. \\ \hline
           external factors & 4 options & to examine the impact of integrating external factors. The range includes the following options (None, USDX, SENT, and USDX + SENT) \\ \hline
        \end{tabular}
    \end{table}
    
    The hyperparameters and their respective ranges (bounds or search space) are as shown in Table \ref{tab:hyperparameters-config}.

        To reduce the problem dimension for the sake of alleviating resources consumption during the running of the GWO, we selected the most essential hyperparameters. The batch size and epoch number have also been excluded from the GWO problem dimensions. Instead, the batch size has been experimentally set to 16, and the early-stop technique is used to terminate epochs automatically. 
    
    \subsection{Evaluation metrics}
        For each model, the GWO has been run multiple independent runs with an objective function to minimise the Mean square error (MSE), which can be calculated according to the equation (6). 
        However, to evaluate the final ensemble and individual models, we used the following six common evaluation metrics:  
        \begin{itemize}
            \item Mean Absolute Error (MAE) measures the average absolute difference between the predicted and actual values.
            \begin{equation}
                \text{MAE}(y, \hat{y}) = \frac{ \sum_{i}^{n} |y_i - \hat{y}_i| }{n} \tag{5}
            \end{equation}
            \item  Mean Squared Error (MSE) for calculating the average squared difference between the predicted and actual values.
            \begin{equation}
                \text{MSE}(y, \hat{y}) = \frac{\sum_{i}^{n} (y_i - \hat{y}_i)^2}{N} \tag{6}
            \end{equation}
            
            \item  Root Mean Squared Error (RMSE), providing a measure of the average magnitude of the error.
            \begin{equation}
                \text{RMSE}(y,\hat{y}) = \sqrt{\frac{\sum_{i}^{n} (y_i - \hat{y}_i)^2}{n}} \tag{7}
            \end{equation}
        \item Mean Squared Prediction Error (MSPE) is a statistical measure used to evaluate the accuracy of a predictive model. It quantifies the average squared difference between the predicted values and the actual observed values. 
        \begin{equation}
        MSPE(y,\hat{y}) = \frac{1}{n} \sum_{i}^{n} (y_i - \hat y_i)^2 \tag{8}
       \end{equation}
        \item Coefficient of Determination ($R^2$ or R-squared) is a statistical measure of how well the regression predictions approximate the actual data points. It is given by the following equation: 
        \begin{equation}
           R^2 (y,\hat{y})= 1 - \frac{\sum_{i=1}^{n} (y_i - \hat{y}_i)^2}{\sum_{i=1}^{n} (y_i - \bar{y})^2} \tag{9}
        \end{equation}
        \end{itemize}
        
        These evaluation metrics allow for quantifying the accuracy and precision of the model's predictions. Lower MSE, MAE, MSPE, and RMSE values indicate better performance, While a high $R^2$ value indicates better performance.

\section{Results and discussion}
    In this section, we first present the optimal configurations identified by the GWO algorithm for each model. We then compare the accuracy of the ensemble forecasts with that of the individual model forecasts.
    
    \subsection{Models optimal configurations}

    Table \ref{tab:GWO results} shows the best solutions identified by GWO for each model. The first column shows the MSE value achieved by the model using the corresponding configuration values, as determined for each model by the GWO optimizer, which are displayed in the subsequent columns.
    
    The results obtained from the GWO-optimised five benchmark models, in Table \ref{tab:GWO results}, showed that the best-performing model is based on the Bi-GRU architecture when incorporating the sentimental score (SENT-Bi-GRU), with the best MSE of 0.000137. The corresponding optimal configuration identified in this model is AdamW as the optimiser with a sliding window of 5 lagged steps, 0.003154571 as the learning rate, 0.0399269013 as the dropout value, and 2 hidden units. Integrating the sentiment score index (SENT) generally provides greater accuracy compared to the USDX index. However, combining both factors improves performance exclusively with the encoder-decoder-LSTM architecture. The optimal values for sliding windows are (5, 6, 10, 17). These figures align consistently with typical workdays, as 5 corresponds to the number of working days a week.

 \begin{table}[bt]

    \caption{Best solutions found by GWO (models optimal configuration). TS = time steps, and h is the number of hidden units}
    \label{tab:GWO results}
    \begin{tabular}{@{}p{3.5cm}lp{1cm}llll}
        \hline
        Model &   MSE &   TS &  LR &  Dropout &  h &  Optimiser \\
        \hline
           SENT-Bi-GRU &  0.000137 &   5 &  0.0031	&  0.3992  &  2 &  AdamW \\  
           SENT-Bi-LSTM  & 0.000194  &    17 & 0.0032 & 0.3205 &  12 &  Adam  \\
           SENT-CNN-Bi-LSTM & 0.000280 &  5 &  0.01 &  0.3439 &  8 &  Adagrad \\
           SENT-CNN-Bi-LSTM-Att &   0.000300 &  6 &  0.0045 &  0.4399  & 8  &  Adam \\
           SENT-USD-CNN-Bi-LSTM-Att & 0.000349   & 10 & 0.0043  &  0.3140  &4 &  RMSprop \\
           SENT-encoder-decoder-Bi-LSTM  & 0.000358  & 10 & 0.0060  & 0.2573  & 64  & Adam\\
           SENT-USD-encoder-decoder-LSTM  &  0.000344 & 17 &  0.0080 &  0.2747  & 64 & AdamW \\
        \hline
\end{tabular}
\end{table}

\begin{figure}[h]
    \centering
    \begin{subfigure}[b]{0.24\textwidth}
        \includegraphics[width=\textwidth]{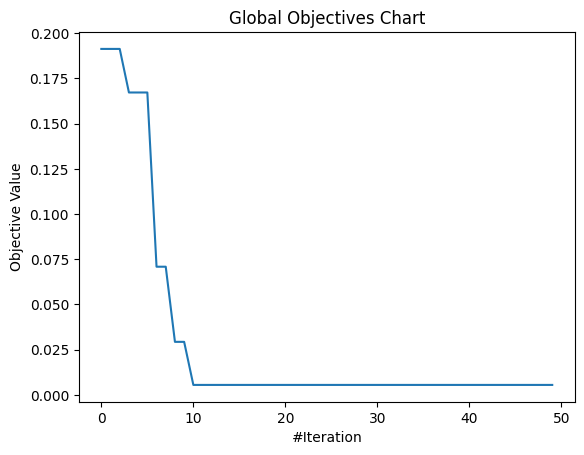}
        \caption{}
    \end{subfigure}
    \begin{subfigure}[b]{0.24\textwidth}
        \includegraphics[width=\textwidth]{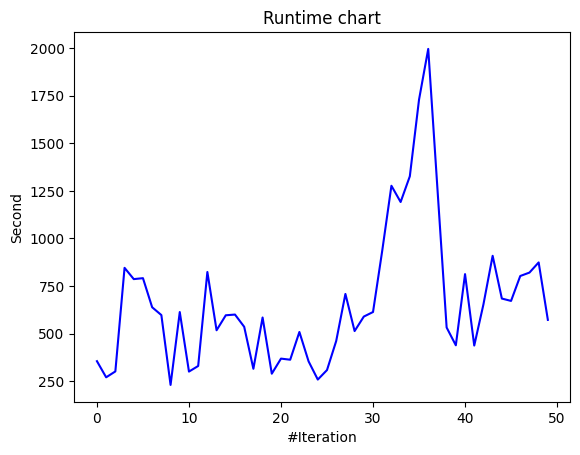}
        \caption{}
    \end{subfigure}
    \begin{subfigure}[b]{0.24\textwidth}
        \includegraphics[width=\textwidth]{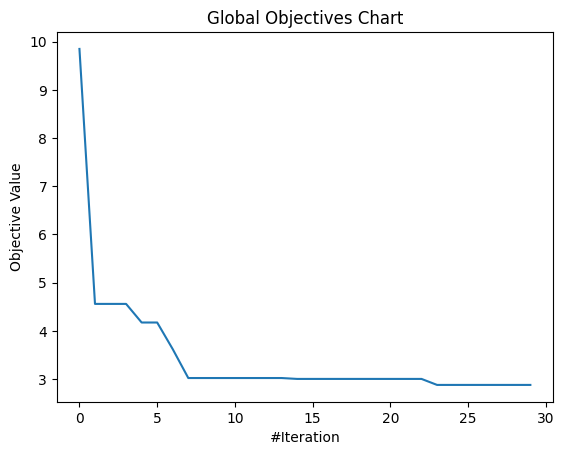}
        \caption{}
    \end{subfigure}
    \begin{subfigure}[b]{0.24\textwidth}
        \includegraphics[width=\textwidth]{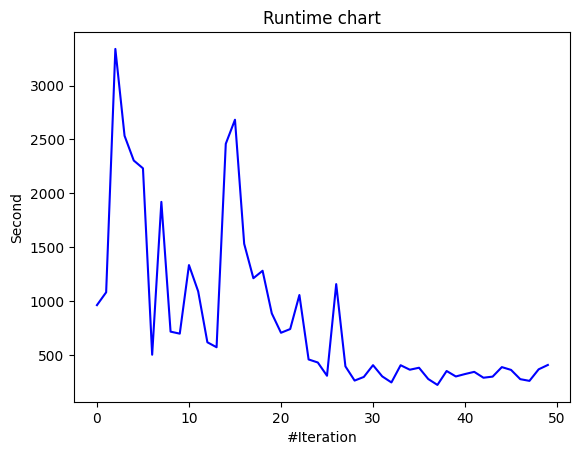}
        \caption{}
    \end{subfigure}
    \begin{subfigure}[b]{0.24\textwidth}
        \includegraphics[width=\textwidth]{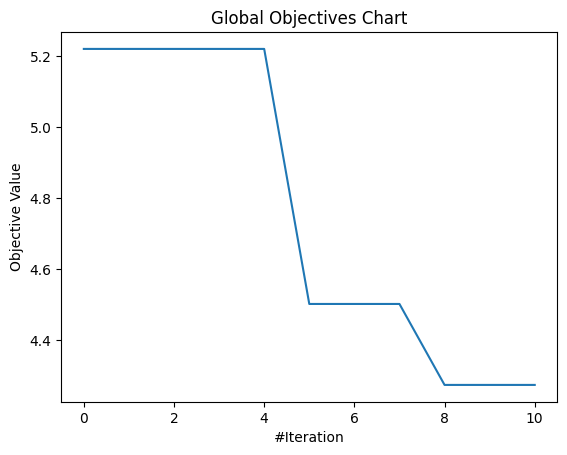}
        \caption{}
    \end{subfigure}
    \begin{subfigure}[b]{0.24\textwidth}
        \includegraphics[width=\textwidth]{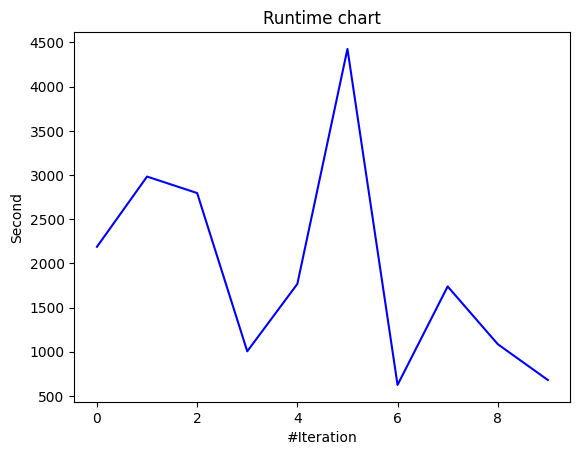}
        \caption{}
    \end{subfigure}
    \begin{subfigure}[b]{0.24\textwidth}
        \includegraphics[width=\textwidth]{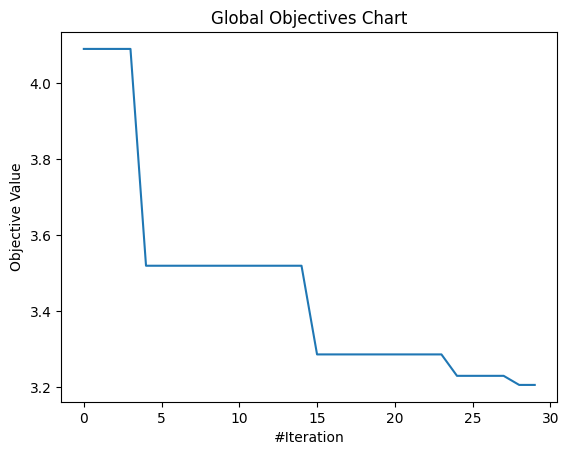}
        \caption{}
    \end{subfigure}
    \begin{subfigure}[b]{0.24\textwidth}
        \includegraphics[width=\textwidth]{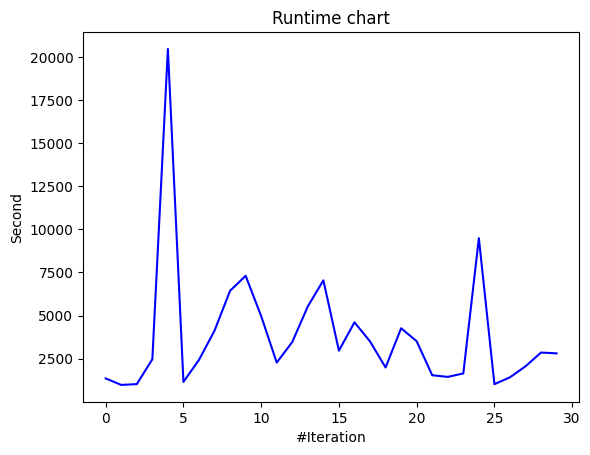}
        \caption{}
    \end{subfigure}
    \caption{Objective function and runtime charts of a single GWO run for various models: (a,b): LSTM, (c,d): GRU, (e,f): CNN-LSTM-att, (g,h): Encoder-Decoder.}
    \label{fig:chartsbasemodels}
\end{figure}

   Conversely, Figure \ref{fig:chartsbasemodels} illustrates global objective and run-time charts. The global objective chart shows the movement curve towards the optimal solution to minimise the MSE for a single model. Like manual hyperparameter tuning, the GWO algorithm takes longer to find a near-optimal solution with the encoder-decoder-LSTM and CNN-LSTM-att models. At the same time, finding a near-optimal solution with LSTM and GRU is faster. Experimentally, GWO reaches a near-optimal solution during the first 30 epochs.

    \subsection{GWO-ensemble model evaluation}

    \begin{table}[h]
    \setlength{\tabcolsep}{3.3pc}
    \caption{GWO-ensemble model optimal weights}
    \label{tab: ensemble model-optimal weights}
        \begin{tabular}{@{}lc}
        
        \hline
           Model     & Optimal weight  \\ \hline
           SENT-Bi-GRU & 0.0860\\ 
           SENT-Bi-LSTM & 0.6266 \\ 
           SENT-Bi-CNN-LSTM-att & 0.0889 \\ 
           SENT-encoder-decoder-LSTM & 0.1057 \\ 
           SENT-USDX-encoder-decoder-LSTM & 0.0927 \\ \hline
        \end{tabular}
    \end{table}
        The experiments showed that this metaheuristic-calibration-based weighted ensemble method achieves better results than traditional method or other naive ensemble methods and that the ensemble model outperforms the individual models, achieving a mean square error of (0.000127) as shown in Table \ref{tab:final evaluation}.
        The best final forecast has been acquired by blending the forecasts obtained from SENT-Bi-GRU, SENT-Bi-LSTM, SENT-Bi-CNN-LSTM-attention, SENT-encoder-decoder-LSTM and SENT-USDX-encoder-decoder-LSTM models. The optimal weights assigned for each model are illustrated in Table \ref{tab: ensemble model-optimal weights}.
\begin{figure}[h]   
    \centering
    \begin{subfigure}[b]{0.32\textwidth}
        \centering
        \includegraphics[width=\textwidth]{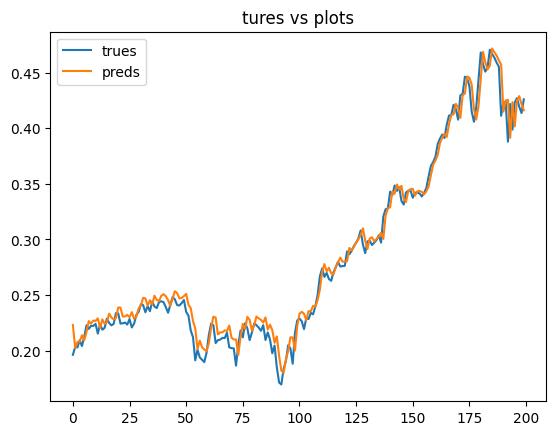}
        \caption{}
    \end{subfigure}
    \begin{subfigure}[b]{0.32\textwidth}
        \centering
        \includegraphics[width=\textwidth]{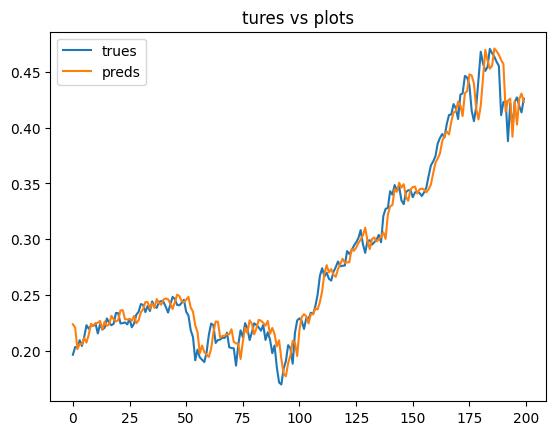}
        \caption{}
    \end{subfigure}
    \begin{subfigure}[b]{0.32\textwidth}
        \centering
        \includegraphics[width=\textwidth]{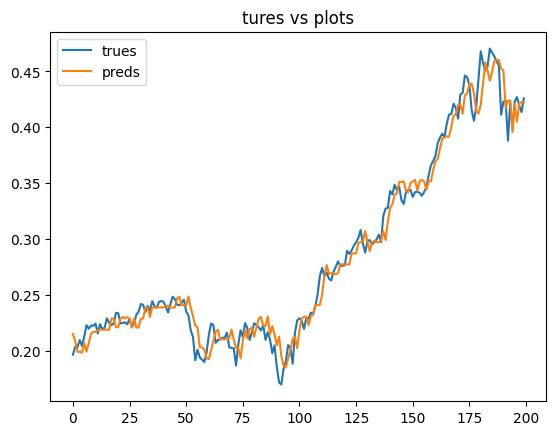}
        \caption{}
    \end{subfigure}

    \begin{subfigure}[b]{0.32\textwidth}
        \centering
        \includegraphics[width=\textwidth]{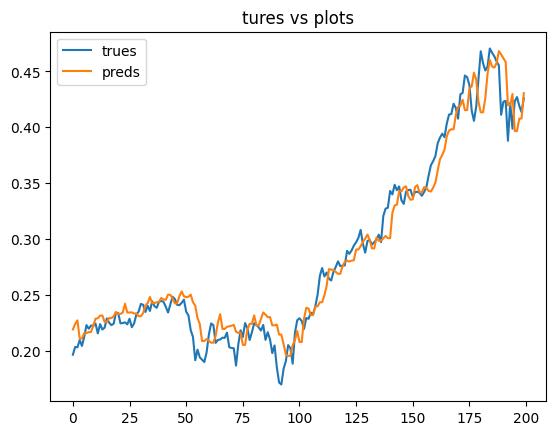}
        \caption{}
    \end{subfigure}
    \begin{subfigure}[b]{0.32\textwidth}
        \centering
        \includegraphics[width=\textwidth]{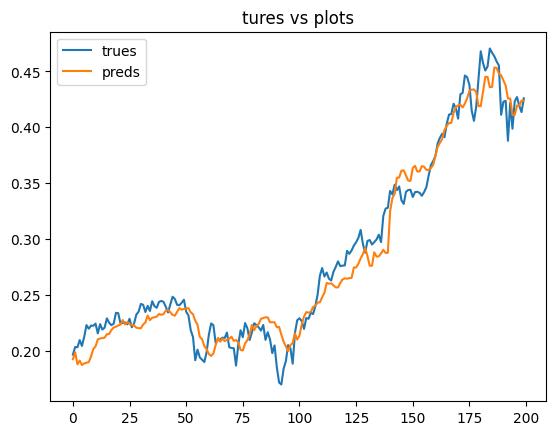}
        \caption{}
    \end{subfigure}
    \begin{subfigure}[b]{0.32\textwidth}
        \centering
        \includegraphics[width=\textwidth]{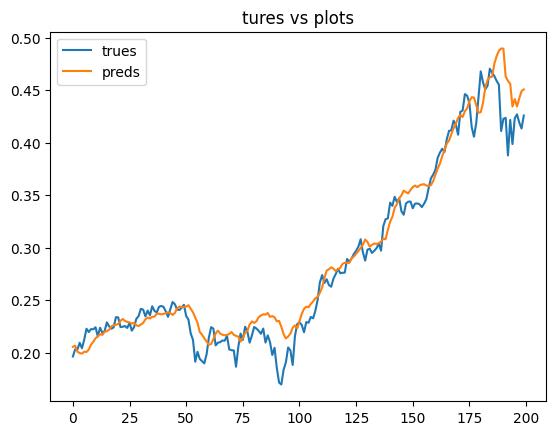}
        \caption{}
    \end{subfigure}
    
    \caption{ Actual vs. predicted results of the examined models: (a). Ensemble model, (b). SENT-Bi-GRU, (c). SENT-Bi-LSTM, (d). SENT-CNN-Bi-LSTM, (e). SENT-CNN-Bi-LSTM-att, (f). SENT-USD-Encoder-Decoder-Bi-LSTM}
    \label{fig:trues vs preds}
\end{figure}
\begin{table}[h]
\caption{Forecasting accuracy evaluation}
\label{tab:final evaluation}
\begin{tabular}{p{4cm}lcccccc} 
\toprule
  Model & MAE & MSE & RMSE & MSPE & MAPE & $R^2$\\
        \hline
        \textbf{GWO-Ensemble } & 0.007414   & \textbf{ 0.000127} & 0.011269  & 0.001649 & 0.026401 & 0.981353 \\
        SENT-BI-GRU & 0.008300 & 0.000137 & 0.011704 & 0.002089 & 0.031694 & 0.980519 \\
         SENT-BI-LSTM & 0.009800 & 0.000194 & 0.013928 & 0.002722 & 0.036012 & 0.972414 \\
         SENT-CNN-BI-LSTM & 0.012650 & 0.000280  & 0.016733 & 0.001633 & 0.031219 & 0.853016\\
         SENT-CNN-BI-LSTM-att & 0.013550 & 0.000300 & 0.017306 & 0.004602 & 0.050919 & 0.957414 \\
         SENT-USD-CNN-BI-LSTM-att & 0.014400& 0.000349 & 0.018682 & 0.006660 & 0.057875 & 0.950375 \\
         SENT-Encoder-Decoder-BI-LSTM & 0.012550 & 0.000359 &   0.018934 & 0.005175 & 0.046804 & 0.949024 \\
         SENT-USD-Encoder-Decoder-BI-LSTM & 0.012900 & 0.000344 &  0.018547 & 0.005353 & 0.049209 & 0.951086 \\
\bottomrule
\end{tabular}
\end{table}

Figure \ref{fig:trues vs preds} illustrates the forecasting performances for all individual models. The figure shows the plot chart of Actual vs. predicted results. Table \ref{tab:final evaluation} shows the overall comparison of the GWO-ensemble model with the other benchmark models.

\clearpage
\section{Limitations}

Although metaheuristic optimisation algorithms effectively tune deep learning models, they require significant computational resources. This limitation has forced us to restrict our hyperparameter search to a limited number of parameters. However, expanding the problem dimension to include a broader range of parameters could improve forecasting performance. Suppose we need to run GWO 20 independent runs for the Bi-LSTM model with 30 epochs and a population size of 10. Each Bi-LSTM model will be trained 6,000 times (20 runs * 30 epochs * 10 population size).

\section{Conclusions}
In this study, we demonstrated how the accuracy of Brent oil price forecasting can be enhanced through the use of ensemble models and advanced optimization techniques. We have introduced a hybrid approach that combines established deep learning networks with Grey Wolf Optimizer (GWO) to enhance the forecasting of Brent crude oil prices. The main idea is to  optimise configurations at various stages of constructing a forecasting solution, alongside the power of combining forecasts with multiple deep learning models. The GWO algorithm was employed to optimise feature selection, dataset preparation, tuning hyperparameters for five individual deep learning networks, and the fusion of their forecasts. The results demonstrated that this approach improved the accuracy for each model and that the presented ensemble model outperforms benchmark models and achieves an MSE of 0.000127.

\section*{Declarations}

\begin{itemize}
\item Funding:
 No external funding was received for this research.
\item Conflict of interest/Competing interests:
 Not applicable.
\item Ethics approval and consent to participate:
 Not applicable.
\item Data availability:
The dataset used in this paper can be accessed through this \href{https://github.com/Med-Rokaimi/Exploratory-data-analysis--Brent-crude-oil} {GitHub repos}
\item Code availability:
 Will be available upon request.
\item  Corresponding author:
  Mohammed Alruqimi
\end{itemize}

\bibliography{ref}

\end{document}